\documentclass[runningheads]{llncs}
\usepackage{graphicx}
\usepackage{amsmath,amssymb,amsfonts}
\usepackage{caption}
\usepackage{subcaption}
\usepackage{hyperref}
\hypersetup{
    colorlinks=true,
    linkcolor=blue,
    filecolor=magenta,      
    urlcolor=blue,
    citecolor=black,
}

\begin{document}
\title{Quadratic GCN for Graph Classification}
%
%
\author{Omer Nagar\inst{} \and
Shoval Frydman\inst{} \and
Ori Hochman\inst{} \and
Yoram Louzoun\inst{}}
\authorrunning{O. Nagar et al.}
%
\institute{Bar-Ilan University, Israel}


\maketitle              
\begin{center}
    {
\email{ovednagar@hotmail.com\\ 123shovalf@gmail.com\\ ori.hochman.work@gmail.com\\ louzouy@math.biu.ac.il}}
\end{center}

\begin{abstract}
Graph Convolutional Networks (GCNs) have been extensively used to classify vertices in graphs and have been shown to outperform other vertex classification methods. GCNs have been extended to graph classification tasks (GCT). In GCT, graphs with different numbers of edges and vertices belong to different classes, and one attempts to predict the graph class. GCN based GCT have mostly used pooling and attention-based models. The accuracy of existing GCT methods is still limited. We here propose a novel solution combining GCN,  methods from knowledge graphs, and a new self-regularized activation function to significantly improve the accuracy of the GCN based GCT. We present quadratic GCN (QGCN) - A GCN formalism with a quadratic layer. Such a layer produces an output with fixed dimensions, independent of the graph vertex number. We applied this method to a wide range of graph classification problems, and show that when using a self regularized activation function, QGCN outperforms the state of the art methods for all graph classification tasks tested with or without external input on each graph. The code for QGCN is available at:\\ \url{https://github.com/Unknown-Data/QGCN}.

\keywords{Machine Learning \and GCN \and Quadratic activation \and graph classification \and activation function}
\end{abstract}

\section{Introduction}
Information is often represented through relations between entities. A standard formalism to represent such relations are graphs, where relations are represented by edges that can be directed and weighted. Such graphs are often used for vertex classification tasks, where one aims to classify the vertices of a given graph. For example, given citations between papers, one could attempt to classify the paper domain ~\cite{lu2003link}. Such tasks are often performed through information diffusion in the network, following the assumption that neighboring vertices have similar classes ~\cite{bhagat2011node,tang2011leveraging,angelova2006graph}. Recently more complex approaches have been proposed to use the network topology ~\cite{rosen2015topological,naaman2018edge,grover2016node2vec} or the propagation of external information ~\cite{kato2009robust,speriosu2011twitter}. A dominant approach for such classifications (either supervised or semi-supervised) is Graph Convolutional Networks (GCN) ~\cite{kipf2016semi}. GCNs are based on weighting a signal propagated through first neighbors in the graph in each layer of the neural network. The general formalism of a layer in GCNs is:
\begin{equation}
X_k=\sigma(X_{k-1}*\tilde{A}*W_k),
\end{equation}
where $A$ is a matrix derived from the adjacency matrix. Often, the symmetric and normalized adjacency matrix in addition to the identity is used:
\begin{equation}
\tilde{A}=D^{-1/2}[A+A^T+I]D^{-1/2}.
\end{equation}
$W_k$ and $X_k$ are weights and input values of the $k^{th}$ layer, where the input to the first layer is the available information on vertices. The output layer has dimensions of the vertex number by the number of classes. 
A large number of extensions and applications of GCN have been published in combination with many other learning methods, including, among others, combinations of GCN with recurrent neural networks ~\cite{ling2019fast}, with GANs ~\cite{lei2019gcn} and with active learning ~\cite{abel2019regional}. 

A related, but much less studied task is the Graph Classification Task (GCT), where each graph belongs to one of $N_C$ different classes. For example, proteins can be classified based on their molecular interaction graphs, organizations can be grouped based on their communication graphs. GCT pose a challenge to GCN based approaches since those produce a projection for each vertex. As such their output dimension is determined by the number of vertices. Current approaches for reducing the output dimension to a fixed value are based on statistical properties of the vertex projection or the network itself ~\cite{pan2015joint,pan2016task}, or pooling ~\cite{zhang2018end}. 
We here propose a direct approach, adopted from knowledge graphs ~\cite{hamilton2017representation} - an additional  quadratic layer to the GCN :
\begin{equation}
X_{end}=\sigma(V_1^T*F(X_{k_i})^T*\tilde{A}*X_k*V_2),
\end{equation}
where $V_1$ and $V_2$ are different weight matrices, and $X_k$ is the input values to the $k$ layer. $F(X_{k_i})$ is some combination of the input to the different layers (e.g. concatenation of the input to different layers). Such a formalism outputs a result of fixed dimensions equal to the external dimensions of $V_1$ and $V_2$. We here propose that using such a formalism with an output vector of the size of the number of classes can be directly used to classify networks. We denote this formalism Quadratic GCN (QGCN). QGCN produces for each network, through a soft-max, a probability for each of the possible classes. We further propose to use as input to the first layer of the QGCN a set of basic topological features, such as the degree, or the number of triangles around each vertex. A limitation of such inputs is that their distribution is often imbalanced, with a large number of vertices with a zero input. We  propose a novel activation function to treat such  input.

\section{Relation to Previous Work}
GCTs have emerged in multiple independent domains, including Graph kernel analysis, GCN,  chemoinformatics, attention-based methods, and self-supervised based methods.

In chemoinformatics, the function of molecules is predicted from their structure and composition. Each molecule is represented as a network of atoms or amino acids, and each vertex is associated with biochemical properties, such as its molecular weight, charge. etc. Edges can also be associated with the physical distance between the vertices, or the type of molecular bond.  We have here tested three datasets of such molecules, including the classification of protein groups ~\cite{borgwardt2005protein}, the classification of mutagens ~\cite{kazius2005derivation} and the classification of molecules that can be used as drugs ~\cite{gauzere2012two}. 

GCTs have been performed using kernel approaches. Graph kernels exploit compare graphs through the frequency of subgraphs. Kernel methods typically restrict themselves to comparing substructures computable in a polynomial time. Many graph kernels have been defined, such as random walks ~\cite{kashima2004kernels,gartner2003graph}, shortest paths ~\cite{assfalg20063dstring}, limited size graphlets (also denoted motifs) ~\cite{shervashidze2011weisfeiler} and subtrees ~\cite{ramon2003expressivity}. ~\cite{Smoothing} proposed a general smoothing framework for graph kernels by taking structural similarity into account, and apply it to derive smoothed variants of popular graph kernels. ~\cite{survey_kernels} gives a full survey of these methods. Note that once a distance between graphs has been defined, the classification itself is straightforward using distance-based classification methods.

GCNs have been also extended to GCT. Important efforts in this direction include DIFFPOOL, a differentiable graph pooling module that can generate hierarchical representations of graphs and use this hierarchy of vertex groups to classify graphs ~\cite{ying2018hierarchical}. Zhang et al. ~\cite{zhang2018end} have used a formalism very similar to the Kipf and Welling formalism but added a pooling strategy to the last layer with a SortPooling layer. Pan et al. have produced a couple of algorithms of Multi-Task Learning (MTL). MTL jointly discovers discriminative subgraph features across different tasks and uses them to classify graphs \cite{pan2015joint,pan2016task}. As such, their work may be seen as an extension of kernel methods. StructPool ~\cite{structpool} considers the graph pooling as a vertex clustering problem, which requires the learning of a cluster assignment matrix. EigenGCN ~\cite{EigenPooling} proposes a pooling operator EigenPooling based on graph Fourier transform, which can utilize the vertex features and local structures during the pooling process. Other pooling include among many others are \cite{Haar,EdgeContractionPooling,HierarchicalGraphPooling,cliquepool}. 

Attention and self-attention based methods have also been extended to GCT. A pooling method based on self-attention has been proposed in ~\cite{SelfAttentionGraphPooling}. Self-attention using graph convolution allows the pooling method to consider both vertex features and graph topology. DAGCN ~\cite{DAGCN} added an attention graph convolution layer to a regular GCN to learn automatically the importance of neighbors at different distances, then employed a second attention component, a self-attention pooling layer, to generalize the graph representation from the various aspects of a matrix graph embedding. ~\cite{GraphStar} is a unified graph neural net architecture that utilizes message-passing relay and attention mechanism for multiple prediction tasks, including GCT. More attention-based methods are \cite{UniversalGraphTransformer,ImprovingAttentionMechanism}.

GCT has also been expanded to self-supervised learning. In many real-world problems, the number
of labeled graphs available for training classification models
is limited, which renders these models prone to overfitting. ~\cite{CSSL} proposed two approaches based on contrastive self-supervised learning (CSSL) to alleviate overfitting. One uses CSSL to pre-train graph encoders on widely-available unlabeled graphs, while in the other a regularizer based on CSSL is developed. Other approaches such as ~\cite{InfoMap2020} and ~\cite{GCC2020} have focused on learning representations of local elements in graphs, such as nodes and subgraphs, but can also be extended to graph classification.

We here propose to adapt from knowledge graphs quadratic layers and merge them with a GCN formalism to classify graphs. A knowledge graph (KG) is a multi-digraph with labeled edges, where the label represents the type of the relationship. Such graphs have been used in multiple GCTs. A general framework for KG is first an encoder to project the vertices, and then a decoder that compresses vertex representations into a single representation. Then, to classify full graphs, Duvenaud et al. ~\cite{duvenaud2015convolutional} averaged all the vertex projections. Gilmer et al. ~\cite{gilmer2017neural}  used a DeepSet aggregation. Li et al. ~\cite{li2015gated} added a new virtual vertex and connected it to all the vertices. They then used the representation of the virtual vertex as the representation of the graph. However, our interest in KG is in the bilinear decoders proposed for link predictions \cite{wang2018multi}. We here propose to merge bilinear decoders with GCN to produce graph classifiers.

\section{Novelty}
The  three main novel aspects of this work are:
\begin{itemize}
\item A quadratic formalism for GCT that is a direct extension of GCNs. This formalism is a GCN on all but the last layer, followed by a bilinear GCN last layer. This formalism produces a fixed size output for graphs of different sizes.
\item A significant improvement in classification accuracy over previously studied  GCT,  either binary or multi-class.
\item A new method to classify graphs in the absence of external information (beyond the adjacency matrix) using topological features of each vertex as an input to each vertex in the GCN.
\item A novel self-regularized activation function to handle inputs that are imbalanced with a large number of zero inputs.
\end{itemize}

\section{Model}
Each vertex is assigned an input vector $x_0(i)$ defined to be either external information on each vertex or topological features of the vertex ~\cite{rosen2015topological}. The values of the following layers are computed as in Eq. 4, with a distinction between the internal layer and the last layer. 
\begin{eqnarray}
\tilde{A}=D^{-1/2}[A+A^T+I]D^{-1/2}\: or \:\tilde{A}=[A+A^T+I] \\ \nonumber
X_k=\sigma_1(X_{k-1}*\tilde{A}*W_k) \quad\quad\quad\quad\quad\quad\\ \nonumber
X_{end}=\sigma(V_1^T*F({X_i})^T*\tilde{A}*X_k*V_2),\quad\quad \nonumber
\end{eqnarray}
The non-linearity in the intermediate layers is:
\begin{equation}
\sigma_1(x)=1 - \frac{2}{x^2+1}.
\end{equation}
This activation function will be further discussed. The activation function in the last layer is $\sigma(x)=\frac{1}{1+e^{-x}}$ for the binary cases, and a softmax otherwise. $F$ is a function of the input to all layers, including the first layer. 

The formalism above translates each graph into a constant size output (Figure 1). As such, it is not sensitive to the graph size. The output is directly amenable to standard loss minimization for graph classification problems by comparing the fixed size output with the expected classification of the graph. Here, a binary cross-entropy loss was used, and an ADAM optimizer.

\begin{figure}[t]
\begin{center}
\centerline{\includegraphics[width=8.5cm, height=4.5cm]{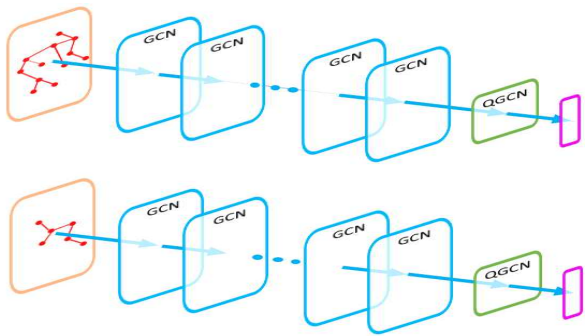}}
\caption{Model description. Graphs of different sizes are passed through GCN layers. The output size of each GCN layer is proportional to the number of vertices in the current graph. The last layer is a quadratic layer with input size proportional to the graph number of vertices and constant size output size. Note that the dimension of the weights of the QGCN are not affected by the graph vertex number, and as such can be learned on all graphs simultaneously.}
\label{QGCN-sturcture}
\end{center}
\end{figure}

The model's input consists of: A) The normalized adjacency matrix of the graph  $\tilde{A}=D^{-0.5}*[A+A^T+I]*D^{-0.5}$. We have tested other normalization methods for the adjacency, but all led to significantly lower accuracy in the training sets. B) A topological input matrix $x_0 \in R^{n*d}$ for each graph, where $n$ is the number of vertices in the graph.  The features used were: In/Out degree (or just degree if the graph is undirected), Centrality and first and second moments of the distance distribution of each vertex to all other vertices in the graph ~\cite{benami2019topological}. When supplied with the graphs, an embedding vector for each vertex was used, instead of the topology.

\section{Implementation}
We compared the accuracy of QGCN using different datasets, as measured by accuracy score. Each dataset was shuffled and split according to the split ratio of the compared experiments. The weights were trained to minimize the loss function over all studied networks in the training set, with equal weights to all networks, independently of their vertex number. 

Minimal hyper-parameter tuning was performed.  The regularization was a weak L2-regularization with coefficients of [1e-7, 1.e-8, 1.e-9, or 0]. The normalization of the input was either z-scoring or standardization of the input to the $0-1$ (further denoted Min-Max normalization). Two version of the normalized adjacency matrix $\tilde{A}$ were used, as in equation 4, and the input features to be used (see details in Table 2). Finally batch sizes between $32$ and $128$ in jumps of $2$ were tested. The test set was not used for any of the parameter tuning steps (See Table 2 for hyper-parameters).

\subsection{Datasets Studied}
We studied multiple previously tested datasets and compared our results on those to the state of the art GCT  algorithms. The datasets properties are summarized in Table 1. 

{\bf HIV (Aids) related molecules ~\cite{gauzere2012two}}. Each graph represents a molecular compound. Each compound is classified as active or inactive. The labels indicate whether a molecular compound is active against HIV or not. The vertices and the edges of the graphs represent atoms and their covalent bounds respectively. The dataset also contains non-topological information about the vertices of the graph such as the chemical compound and the charge of the atom.

{\bf Mutagens ~\cite{kazius2005derivation}}. As in the HIV dataset,  molecules are converted to graphs, and each graph is labeled as mutagen or non-mutagen. 

{\bf Proteins classification Dataset ~\cite{protein03}}. PROTEINS is a protein graph dataset, where vertices represent the amino acids and two vertices are connected by an edge if they are less than 6 Angstroms apart. The label indicates whether or not a protein is an enzyme.

{\bf Grec drawing classification. ~\cite{nene96columbia}}. The dataset consists of graphs that were produced by symbols from architectural and electronics drawings. The images occur in five different distortion levels and the graphs are extracted from the resulting de-noised images by tracing the lines from end to end and detecting intersections as well as corners. the dataset contains 1,100 graphs uniformly distributed over 22 classes.

{\bf NCI1 and NCI109 ~\cite{nci1nci109}} are two balanced biological datasets screened for activity against non-small cell lung cancer and ovarian cancer cell lines, where each graph is a chemical compound with vertices and edges representing atoms and chemical bonds, respectively. 

The HIV, Mutagens, and Grec datasets contain information about each vertex, while Proteins, NCI1, and NCI109 do not.

\begin{table}[t]
\centering
\caption{Datasets Statistics}\label{Datasets-Statistics}
\begin{tabular}{lcccc}
\hline
Dataset & Num graphs & Avg vertices & Avg edges & Classes\\
\hline
Mutagencity & 188 & 17.93 & 19.79 & 2 \\ 
NCI109 & 4127 & 29.68 & 32.13 & 2 \\
NCI1 & 4110 & 29.87 & 32.30 & 2 \\
Grec & 1100 & - & - & 22 \\
Protein & 1113 & 39.06 & 72.82 & 2 \\
Aids & 2000 & 15.69 & 16.20 & 2 \\
\hline
\end{tabular}
\end{table}

\subsection{Batch Optimization of QGCN}
In order to run QGCN in a parallel batch form to allow for GPU based processing, one must have input of equal size. While the weights matrices are not affected by the graphs number of edges and vertices, the matrix product of $\tilde{A}$ and $X_k$ is, preventing a tensor representation of the loss computation and the resulting back propagation. To improve batch based computation,  QGCN  uses a padding technique that allows it to stack the $\tilde{A}$ of all graphs into one single tensor that has no effect on the outcome.  Zero padding from the right (and bottom) of both $\tilde{A}$ and $X_k$ has no effect on the resulting $\tilde{A}X_k$ computation for all layers, except for changing the output size. However, since the last layer output size is constant, this is of no importance. We thus compute for each batch the maximal vertex number, and zero pad the other graphs to this size. At the current stage, QGCN does not handle sparse matrices.

\begin{table}[t]
\centering
\caption{Fine-tuned hyperparamters. For all datasets: train-validation-test split is 0.675-0.125-0.2, learning rate is 1e-4, an Adam optimizer is used, 2 layers of GCN in size 250 each are applied and dropout=0. The hyperparameters that are different for each dataset are summarized here. NR corresponds to Norm reduced normalization of the adjacency matrix, while NRS is the symmetric normalization. In features, B is BFS, C is centrality and D is degree.}\label{Finetuned hyperparameters}
\begin{tabular}{lcccccc}
\hline
Hyperparameter& Mutagencity & NCI109 & NCI1 & Grec & Proteins & Aids\\
\hline
Batch size & 128 & 128  & 32& 32  & 128 & 128 \\
L2 regularization & 1e-7 & 1e-9  & 1e-7& 1e-9  & 1e-9 & 0 \\
Standardization & z-score & z-score  & min-max & min-max  & min-max & z-score \\
Adjacency norm & NR & NR  & NRS & NRS  & NRS & NR \\
Features & C, D & B, C, D  & B, C, D & B, C, D  & B, C, D & B, C, D \\
\hline
\end{tabular}
\end{table}

\section{Results}
We tested the accuracy of the QGCN on the datasets above, as measured by the test accuracy and compared it to several baselines that can be divided into four groups:
{\bf Graph Kernel Methods.} Shortest-Path Kernel (SP) ~\cite{SP}, Weisfeiler-Lehman Kernel (WL) ~\cite{nci1nci109} and GRAPHLET ~\cite{GRAPHLET}. {\bf Graph Pooling Models.} Models that combine GNNs with pooling operator for graph level representation learning: HaarPool ~\cite{Haar}, HPG-SL ~\cite{HierarchicalGraphPooling}, G-Inception ~\cite{Ginception}, WKPI ~\cite{wkpi}, EigenGCN ~\cite{EigenPooling} and DGCNN ~\cite{DGCNN}.{\bf Attention based methods}. DAGCN ~\cite{DAGCN}, SAG-Pool ~\cite{SelfAttentionGraphPooling} and GAT-GC ~\cite{gatgc}. {\bf Self-supervised Methods.} CSSL ~\cite{CSSL}.

Table 3 presents the accuracy of QGCN on graph classification task compared to the above baselines. Here, the quadratic layer (last layer of the GCN) is calculated with $F(X_{k_{i}})=X_0$ where $X_0$ is the input layer of the GCN.
We have performed for each training set fraction, 20 trainings of the model (for the same training/validation/test division and the same parameters), and used the one providing the highest accuracy on the validation set. In all datasets, both binary and multi-binary, QGCN reached a much higher test-set accuracy than all previously published algorithms, with an improvement that can reach $0.05$ vs the top competing algorithm. In the Aids (HIV) reaction and Grec, we even obtain a prefect classification accuracy. The results are not very sensitive to the hyperparameters used, with three exceptions. The internal non-linearity function is crucial, a low regularization must be used, and the $F(X_{k_{i}})$ should simply be the input to the graph.

Table 4 shows QGCN performance on GCT where $F(X_{k_{i}})=C$ where $C$ is the concatenation of the input to the different layers, or $F(X_{k_{i}})=X_{k-1}$ where $X_{k-1}$ is the last layer of the GCN (hyperparameters are described in Table 2), or as is used in the tables above $F(X_{k_{i}})=X_{0}$. The performance of QGCN is maximal when using the input layer as $F(X_{k_{i}})$. For Mutagencity, the results are quite similar when one combines all input layers, while for the other datasets they are lower. Furthermore, when using a fully quadratic layer, the accuracy is even much lower. Note that more complex functions, such as polynomials of the same matrices could be used. However, here the simplest input layer seems to be the best choice.

\begin{table*}[t]
\caption{Accuracy comparison for graph classification tasks (accuracy score). In this and all tables, the standard error is the standard error on the test on 20 train-validation splits. The result presented is the test accuracy on the run with the best validation accuracy. The same holds for all following tables.}
\label{GCT result}
\begin{center}
\begin{small}
\centering
\scriptsize
\begin{tabular}{cccccccc}
\hline
Categories & Method & Mutagencity & NCI109 & NCI1 & Grec & Proteins & Aids\\ 
\hline
&SP & 80.10$\pm$0.20 &  77.60$\pm$0.30 & 79.30$\pm$0.40& - &75.90$\pm$0.40 & 99.70$\pm$0.00\\
Kernels&WL & 84.50$\pm$0.30 &  86.00$\pm$0.30 & 86.20$\pm$0.10& - &75.50$\pm$0.3 & 99.70$\pm$0.00\\
 &GRAPHLET & 56.65$\pm$1.74 &  60.96$\pm$2.37 &62.48$\pm$2.11& - &72.23$\pm$4.49 & -\\
\hline
&HGP-SL & 82.15$\pm$0.58 &  80.67$\pm$1.16 & 78.45$\pm$0.77& - &84.91$\pm$1.62 & -\\
&G Inception & 95.00$\pm$4.61 & 80.32$\pm$1.73 & -& -& -& -\\
Pooling&WKPI & 88.30$\pm$2.60 & 87.40$\pm$0.3 & 87.50$\pm$0.50 & -& 78.50$\pm$0.40 &- \\
&HaarPool & 90.00$\pm$3.60 &75.60$\pm$1.20 &78.60$\pm$0.50 &- & 80.40$\pm$1.80&- \\
&DGCNN & 80.41$\pm$1.02 & 78.23$\pm$1.31&74.08$\pm$2.19 &- & 79.99$\pm$0.44&- \\
 &EigenGCN &79.50 & 74.90& 77.00& -& 76.60 &- \\
 \hline
&SAGPool &- & 74.06$\pm$0.78 & 74.18$\pm$1.20 & -& 71.86$\pm$0.97 &- \\
Attention&GAT-GC &90.44$\pm$6.44 & - & 82.28$\pm$1.81 & -& 76.81$\pm$3.77 &- \\
 &DAGCN &87.22 & 81.46 & 81.68 & -& 76.33 &- \\
 \hline
Self-Supervised&CSSL &82.64$\pm$0.83 & 81.16$\pm$1.42 & 80.09$\pm$1.07 & -& 85.80$\pm$1.01 &- \\
\hline
\textbf{Proposed} &\textbf{QGCN} & \textbf{95.57$\pm$0.02}& \textbf{90.28$\pm$0.09} & \textbf{91.81$\pm$0.03}& \textbf{100.00$\pm$0.00}& \textbf{87.23$\pm$0.06} & \textbf{100.00$\pm$0.00}\\
\hline
\end{tabular}
\end{small}
\end{center}
\end{table*}

\begin{table}[t]
\caption{Accuracy of QGCN on GCT with different $F(X_{k_{i}})$}
\label{GCT result with different f}
\begin{center}
\begin{small}
\centering
\begin{tabular}{lcccccr}
\hline
$f(X_{k_{i}})$& Mutagencity & NCI109 & NCI1 & Grec & Proteins & Aids\\ 
\hline
$C$ & 93.30$\pm$0.05 & 66.18$\pm$0.07  & 71.58$\pm$0.09&100.00$\pm$0.00&70.55$\pm$0.09 & 100.00$\pm$0.00 \\
$X_{k-1}$ & 90.50$\pm$0.02 &  60.97$\pm$0.09 &65.89$\pm$0.08 & 100.00$\pm$0.00 & 63.77$\pm$0.08 &100.00$\pm$0.00 \\
$X_{0}$ & 95.57$\pm$0.02 &  90.28$\pm$0.09 &91.81$\pm$0.03 & 100.00$\pm$0.00 & 87.23$\pm$0.06 &100.00$\pm$0.00 \\
\hline
\end{tabular}
\end{small}
\end{center}
\end{table}

\section{Self-Regularized Symetric Sigmoid - SRSS}
A problem arising in the context of QGCN is that the input to many vertices is 0. Thus a dual input distribution often emerges, with a large fraction of zero inputs, and a continuous distribution for other vertices. This happens, since for many vertices, either the topological input or the external features input is null. Moreover, when further multiplied by the adjacency matrix, a position with a very low degree can obtain a value of 0 for the input from its neighbors (Figure 2A). Thus, it may be more important to distinguish between zero and non-zero values, than differentiating between positive and negative values. We thus propose a non-linearity that has a minimum at zero to distinguish between zero and non-zero values. We use this function for the QGCN inner layers (the regular GCN layers). In classical activation functions, such as Relu and Tanh, zero has no special value. We propose a continuous smooth differentiable  function:
\begin{equation}
f(x)=G(x)-\frac{2}{x^2+1},
\end{equation}
where $G(x)$ is any existing  activation function, and $x$ is the input value. Figure 2B shows the function graph with $G(x)=1$, as used here. This function is equivalent to:
\begin{equation}
\sigma(log|X|);
\sigma(x) =\frac{e^x-e^{-x}} {e^x+e^{-x}}.
\end{equation}
Similar functions were studied in the context of self-regularized non-monotonic activation function ~\cite{mish}. A general conclusion obtained for functions with non-monotonic nonlinearities at 0 is that such functions may act as a regularizer. Therefore, we propose to use very limited regularization if any.
 
\begin{figure}[th!]
\begin{center}
\begin{subfigure}[b]{0.48\textwidth}
\centerline{\includegraphics[width=\columnwidth]{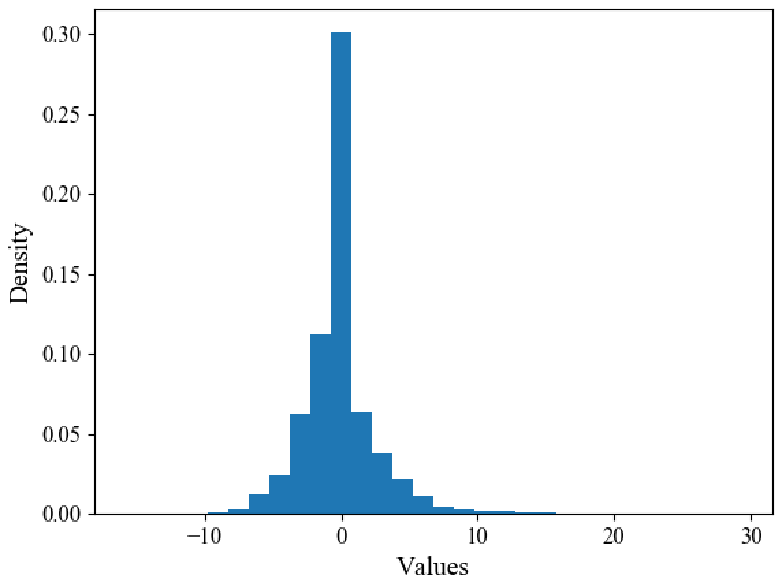}}
\caption{}
\label{Zero-Values}
\end{subfigure}
\begin{subfigure}[b]{0.48\textwidth}
\centerline{\includegraphics[width=\columnwidth]{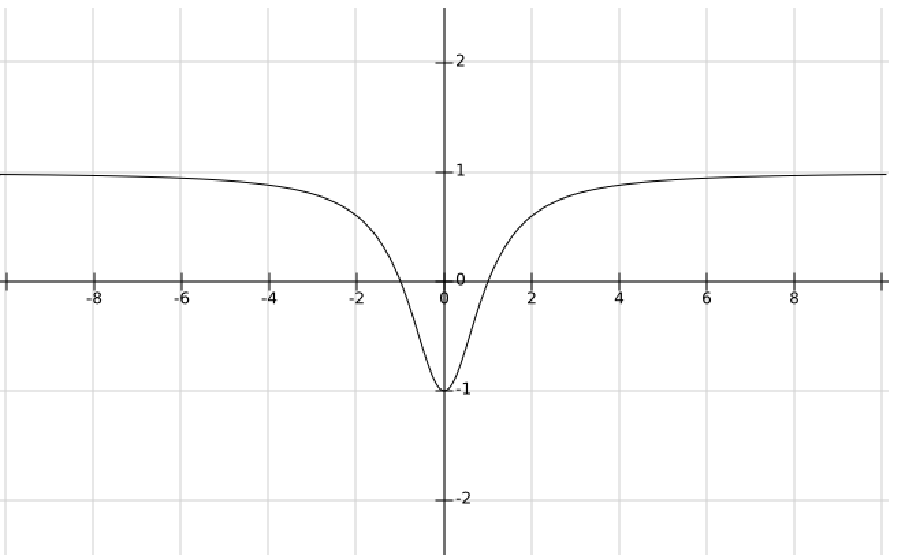}}
\caption{}
\label{SRSS with G(X)=1}
\end{subfigure}
\end{center}
\label{SRSS}
\caption{The first figure is a histogram of the seventh dimension of the output of the first layer of GCN. It can be seen that most values are zero or close to zero. The figure beneath it presents SRSS with G(X)=1}
\end{figure}

\noindent Table 5 shows the Accuracy of QGCN in graph classification task with SRSS against known activation functions such as Relu, Sigmoid, and Tanh. One can see that using SRSS in the inner layers (i.e. in the GCN layers) improves significantly the accuracy of the results. The accuracy of QGCN with other non-linearity is significantly below the accuracy of existing methods.

\begin{table}[t]
\caption{QGCN performance in GCT, with different activation functions (accuracy score)}
\label{GCT different activation functions with QGCN}
\begin{center}
\begin{small}
\centering
\begin{tabular}{ccccccc}
\hline
Activation Function & Mutagencity & NCI109 & NCI1 & Grec & Proteins & Aids\\ 
\hline
Relu & 55.07 & 50.91  & 52.01&  97.29& 74.90 &80.00 \\
Tanh &  57.72& 50.09 & 50.12 &  92.09 & 69.51& 99.72\\
Sigmoid & 64.17 & 52.12 & 62.09 &  90.50 & 73.54 & 99.30\\
\hline
SRSS & {\bf 95.57}& {\bf 93.28} & {\bf 94.81}& {\bf 100.00}& {\bf 87.23} & {\bf 100.00}\\
\hline
\end{tabular}
\end{small}
\end{center}
\end{table}

The bimodal input distribution is also frequent in regular GCN. To further explore the applicability of SRSS, beyond QGCN, we tested on vertex classification task in GCN. ~\cite{ssp} is currently the state-of-the-art GCN which yields the best performance on vertex classification task on known citation datasets such as Citeseer. They use Relu as their activation function. To compare known activation functions and SRSS, we have repeated the experiments in ~\cite{ssp} for each activation function for three citation datasets: Pubmed, Citeseer, and Cora, using the reported hyperparameters. Table 6 shows the results for different activation functions. SRSS yields the same performance as the currently used non-linearities. While the focus here is on GCT, SRSS may be an important tool for other cases with a distinction between the zero values and all other values.

\begin{table}[t]
\caption{GCN performance in node classification task, with different activation functions}
\label{GCT different activation functions with GCN}
\begin{center}
\begin{small}
\centering
\begin{tabular}{ccccccc}
\hline
Activation Function & Cora & Citeseer & Pubmed \\ 
\hline
Relu & {\bf 90.16} &  80.11 &87.84  \\
Tanh & 88.48 &  78.91 & {\bf 88.57}  \\
Sigmoid & 89.86 &  76.70 &87.02  \\
\hline
SRSS & SRSS & 89.54& {\bf 80.52} & 88.49\\
\hline
\end{tabular}
\end{small}
\end{center}
\end{table}

\section{Discussion}
Methods to compare graphs have been developed in many domains, including chemoinformatics  \cite{kashima2004kernels,gartner2003graph}, sub-graph isomorphism ~\cite{shervashidze2009efficient}, graph classification tasks ~\cite{li2015gated} and many others. Many of these methods were kernel-based, where properties of the graph were defined and graphs were compared based on the similarity between graph properties. A simple example would be the frequency of specific subgraphs. An alternative approach is to project the graph and compare the properties of the projection.  Generally, all such methods define a distance matrix between graphs, based on a constant sized representation of the graph. Once such a distance is defined, supervised and unsupervised distance-based approaches can be applied to a set of graphs. 

However, such approaches require a pre-definition of the features to be compared. As such, they require adaptation to different types of graphs. In vertex classification tasks, GCNs have been shown to obtain better accuracies than tailored methods. We here proposed to extend the same logic to GCT. GCN can be viewed as a projection of vertices into a real matrix whose dimensions are the number of vertices and a constant representation dimension. Previous extensions of GCN to graph classification have been proposed based on either pooling methods or through hierarchical approaches. We here suggested an alternative approach based on a quadratic layer, and have shown that combined with a self-regularized activation function, it produced much higher accuracy values than the state of the art methods on all tested GCT.

Instead of pre-defining the properties of graph projections to best classify them, these properties are learned through a combination of weight multiplying different projections of the graphs using a quadratic product. Such products have been previously proposed in multiple contexts, but mainly knowledge graphs. Its extension to regression problems or semi-supervised approaches is straightforward. 

The QGCN can be considered as an attention mechanism on the graph vertices, where one layer servers as the attention and one layer as the projection. Interestingly, the layer producing the highest accuracy is simply the input layer (multiplied by a weight matrix that can be treated as the directions of interest). As such, one may consider choosing an input layer optimized for such attention.

Beyond its usage in the context of GCN, the SRSS function proposed here can have a wide range of applications with bimodal input distributions, where it may be more important to distinguish zero from non-zero-values than positive from negative values. We have here shown that in QGCN, it drastically improves performance and that it can also be used in GCN. We propose this self regularizing function as a basic building tool, at least in the input layer when zero and non-zero values should be separated. 

\bibliographystyle{splncs04}
\bibliography{QGCN.bib}

\end{document}